\documentclass[conference]{IEEEtran}
\IEEEoverridecommandlockouts
\usepackage[T1]{fontenc}
\usepackage{cite}
\usepackage{booktabs}
\usepackage{amsmath,amssymb,amsfonts}
\usepackage{graphicx}
\usepackage{textcomp}
\usepackage{xcolor}
\usepackage{hyperref}
\def\BibTeX{{\rm B\kern-.05em{\sc i\kern-.025em b}\kern-.08em
    T\kern-.1667em\lower.7ex\hbox{E}\kern-.125emX}}
    
\usepackage{breqn}
\usepackage{algorithm}
\usepackage{algpseudocode}
\usepackage{pifont}
\usepackage{listings}
\usepackage{float}
\usepackage{adjustbox}
\def\BibTeX{{\rm B\kern-.05em{\sc i\kern-.025em b}\kern-.08em
    T\kern-.1667em\lower.7ex\hbox{E}\kern-.125emX}}
    
\begin{document}

\title{ChatGPT for Us: Preserving Data Privacy in ChatGPT via Dialogue Text Ambiguation to Expand Mental Health Care Delivery}

%Preserving Data Privacy via Dialogue Text Ambiguation for Employing ChatGPT for Psychological Stress Reduction

%Enabling ChatGPT for Psychological Stress Reduction While Preserving Data Privacy via Dialogue Text Ambiguation

\author{A. Ovalle, M. Beikzadeh, S. Fazeli, P. Teimouri, K.W. Chang, and M. Sarrafzadeh% <-this % stops a space #$^{1}$
\thanks{All authors are with UCLA. \{anaelia,sfazeli,kwchang,majid\}@cs.ucla.edu,}
\thanks{\{mehrabbzapril,parshanteimouri\}@gmail.com.}
}

\maketitle

\vspace{-4cm} 
% rapid reponse, enable rapid responses, enable rapid protocol design to minimize the fallout, 
\begin{abstract}
Large language models have been useful in expanding mental health care delivery. ChatGPT, in particular, has gained popularity for its ability to generate human-like dialogue. However, industries with data-sensitive regulation, such as healthcare, face challenges in using ChatGPT due to privacy and data-ownership concerns. To enable its utilization, we propose a text ambiguation framework that preserves user privacy. We ground this in the task of addressing stress prompted by user-provided texts to demonstrate the viability and helpfulness of privacy-preserved generations and find that recommendations are able to be moderately helpful and relevant, even if original user text is not used. 
\end{abstract}  

\begin{relevance}
This establishes a mechanism for how to use ChatGPT in data-sensitive health applications while preserving data-privacy.
\end{relevance}

\vspace{-0.1cm} 
\section{Introduction}
% \vspace{-0.1cm} 
% Language technologies can help improve mental health outcomes, with abundant scholarly literature indicating utility in AI-augmented assistance to mental health care delivery and interventions more broadly \cite{d2017artificial, de2022augmented, ovalle2021leveraging}. ChatGPT, since its release no less than one year ago, has already disrupted several domains due to its human-like dialogue capabilities. However, towards operationalizing chatGPT, the medical domain necessarily requires abiding by techniques which preserve the privacy of patients\cite{jeblick2022chatgpt, dahmen2023artificial, sallam2023chatgpt}. In this work, we proceed with a text ambiguation framework to help enable the use of ChatGPT in data-sensitive domains and ground it in task centered on reducing stress associated with respect to user-provided texts reflecting economic instability 

Language technologies have proven useful in improving mental health outcomes according to scholarly literature \cite{d2017artificial, ovalle2021leveraging}. ChatGPT has disrupted several domains with its human-like dialogue capabilities, but the health domain requires privacy-preserving techniques. Therefore, this work proposes a text ambiguation framework to enable the use of ChatGPT in data-sensitive domains, focusing on reducing stress related to economic instability\footnote{https://health.gov/healthypeople/priority-areas/social-determinants-health}.

% \vspace{-0.1cm} 
% khan2023chatgpt, lee2023rise

\vspace{-0.25cm} 
\section{Methods}
\vspace{-0.15cm} 
%lets say we define safe as in when data privacy is preserved 
We propose an interactive and privacy-preserving ambiguation framework for text-based recommendations. The framework takes user input (e.g. texts, mobile diary) to populate a masked query with relevant details before eliciting a recommendation from ChatGPT. Illustrated in Figure 1, ChatGPT is provided a masked dialogue question (MDQ), filled by inferred subject matter for the context and therefore - most importantly- does \textbf{not} directly pass any user data. Moving forward, P and NP describe MDQs filled with either the inferred social context or original user text, respectively. In order to test how well our ambiguation framework produces recommendations, we use a subset of the Dreaddit dataset\footnote{http://www.cs.columbia.edu/~eturcan/data/dreaddit.zip} consisting of posts describing economic instability, food insecurity, and housing insecurity (N=110). Each data category is used to fill the MDQ’s context. We evaluate each recommendation across a counselor trainee rubric\footnote{https://www.utoledo.edu/hhs/counselor-education}. Accordingly, we assess positive relationship building, relevance, practicality, and overall perceived helpfulness of the NP responses (Likert Scale, 1-5). We also assess text similarity across NP and P recommendations.

\vspace{-0.1cm} 

% Stress is inextricablytied to social determinants of health (SDOH) such as economic instability, food insecurity, and housing insecurity (cite).

% In order to do this, extracts abstractive context tags (todo:update) from the text  proceed
% Ambiguation framework.....

% give some information about how they're doing... 
% map info to context --> (1) emotion (2) if it is negative/positive/neutral, polarity? --> masked questions --> how do you overcome interpersonal stress? -> chat gpt response

% we will assume for modeling sake that we are trying to maximize something... to do this, help overcome poor. distress tolerance? rumination? 

% if its negative: 
% how does one overcome <neg/pos> <emotion> ? 

% If its just rumination --> then arent i always just giving the same recommendation if i predict it?

\begin{figure}[t!]
    \centering
    \includegraphics[width=0.49\textwidth]{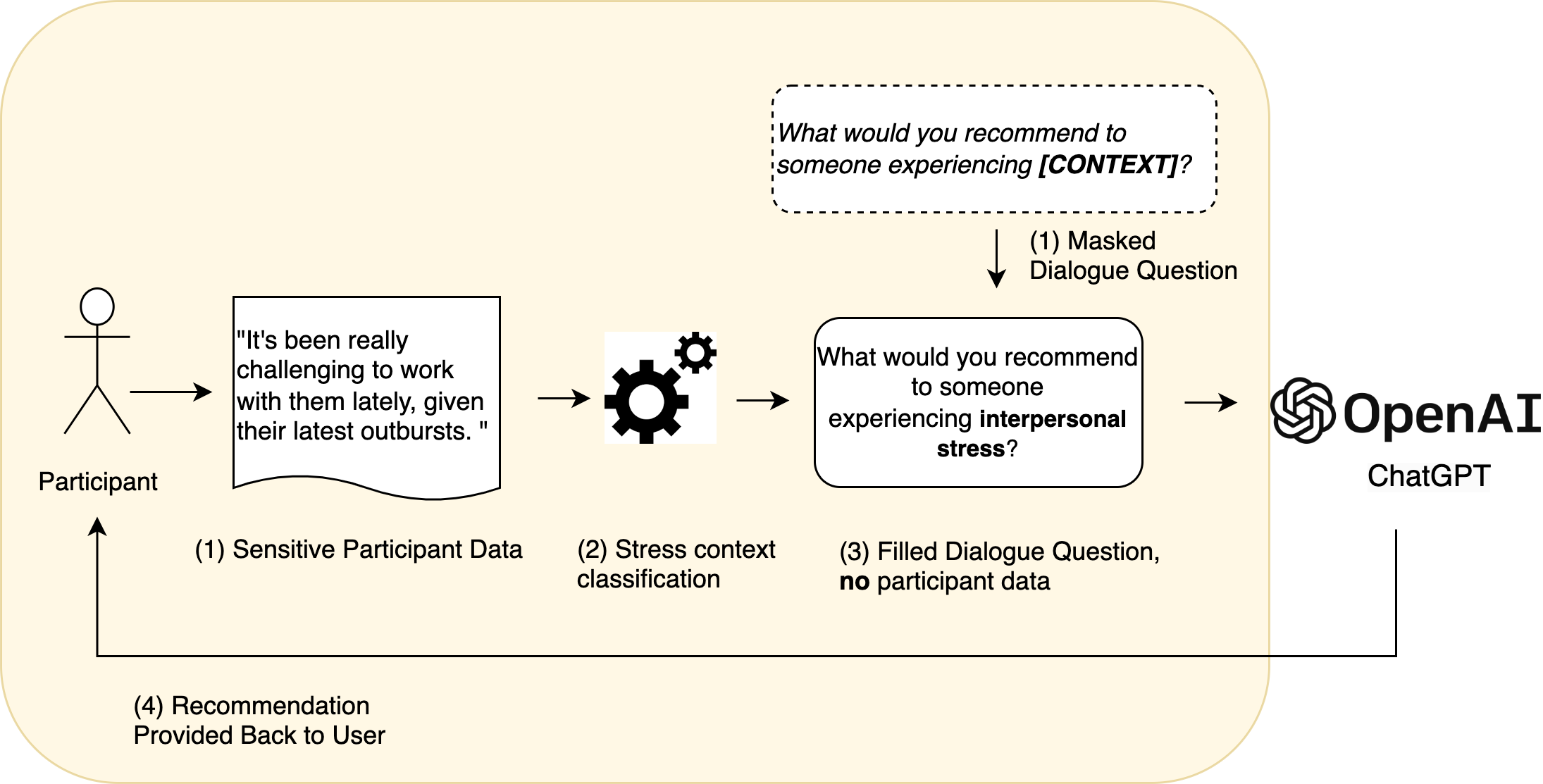}
    \vspace{-0.5cm}
    \caption{Our framework for utilizing ChatGPT while preserving user-privacy.}
    \label{fig:model}
    \vspace{-0.6cm} 
\end{figure}

% \vspace{-0.3cm} 
\section{Results \& Discussion}
\vspace{-0.1cm} 
We measured cosine similarity after calculating TF-IDF on P versus NP responses and found an average score of 0.25, indicating some similarity between responses. Upon review, we observed higher levels of positive relationship building and nuance in NP responses. For instance, NP responses included language that validated and reflected feelings back to the user, as well as expressed more nuanced empathetic statements (e.g. P: ``I'm sorry you're experiencing stress from housing instability'', NP:``It must be really hard to experience this.''). As expected, NP messages are organically more expressive than P messages. We assessed our examples among 3 author annotators and discovered that ChatGPT responses contained pertinent information related to the original Reddit post, even though the original message was not given to ChatGPT (mean Krippendorf $\alpha$=0.30). The scores for positive relationship building, relevance, practicality, and helpfulness averaged to 3.78, 2.69, 2.52, and 2.41. Despite differences in P vs NP inputs, our findings suggest that ChatGPT may still be helpful even when not provided the original user text. We plan to expand how the context is inferred, although this serves as a good starting point to determine the helpfulness of NP recommendations. We acknowledge that our findings are task-specific and encourage future work across several data-sensitive domains. Nonetheless, this work provides direction in navigating ChatGPT operationalization barriers for mental health applications.

% such as incorporating a data-sensitive abstract summarization framework in a way that does not violate privacy.

%when should we
% only provide a rec when wellbeing in danger. this can be grounded in a framework of obvious health risk behavior (risky sex, drug abuse). point of intervention must be contended. here assumptions are made about what one should do? 

% researchers need to be careful about what is considered pathological here. whats considered normal. here the decision lies in whether or not I should change their emotional wellbeing. 

%limitations
% limitation in mundane resposnes if same questions. though there is no deterministic generation of text due to the nature of auto regressive training of chatgpt. if its just one task , need to think about diversity of response. futhermore, mechanism of detection of an emotion is still placed on our inner model. we intend to extend this to 

% future work should vary question selection and determine how well the question fits the task. 

% unclear how to use this for mental health tracking, which requires passing forms of non ambiguated text to the LLM. and unclear whether doing so is helpful longterm due to the blackbox nature of chatgpt, lack of transparency measures, which adds questions to longterm reliability of this model. Regardless, our work plays a first step in enabling assistance that incorporates this tool in a way that effectively preserves user text privacy.

\vspace{-0.3cm} 
\bibliographystyle{IEEEtran}
\bibliography{root.bib}

\begin{thebibliography}{1}
\providecommand{\url}[1]{#1}
\csname url@samestyle\endcsname
\providecommand{\newblock}{\relax}
\providecommand{\bibinfo}[2]{#2}
\providecommand{\BIBentrySTDinterwordspacing}{\spaceskip=0pt\relax}
\providecommand{\BIBentryALTinterwordstretchfactor}{4}
\providecommand{\BIBentryALTinterwordspacing}{\spaceskip=\fontdimen2\font plus
\BIBentryALTinterwordstretchfactor\fontdimen3\font minus
  \fontdimen4\font\relax}
\providecommand{\BIBforeignlanguage}[2]{{%
\expandafter\ifx\csname l@#1\endcsname\relax
\typeout{** WARNING: IEEEtran.bst: No hyphenation pattern has been}%
\typeout{** loaded for the language `#1'. Using the pattern for}%
\typeout{** the default language instead.}%
\else
\language=\csname l@#1\endcsname
\fi
#2}}
\providecommand{\BIBdecl}{\relax}
\BIBdecl

\bibitem{d2017artificial}
S.~D'alfonso, O.~Santesteban-Echarri, S.~Rice, G.~Wadley, R.~Lederman,
  C.~Miles, J.~Gleeson, and M.~Alvarez-Jimenez, ``Artificial
  intelligence-assisted online social therapy for youth mental health,''
  \emph{Frontiers in psychology}, vol.~8, p. 796, 2017.

\bibitem{ovalle2021leveraging}
A.~Ovalle, O.~Goldstein, M.~Kachuee, E.~S. Wu, C.~Hong, I.~W. Holloway, and
  M.~Sarrafzadeh, ``Leveraging social media activity and machine learning for
  hiv and substance abuse risk assessment: development and validation study,''
  \emph{Journal of Medical Internet Research}, vol.~23, no.~4, p. e22042, 2021.

\end{thebibliography}

\end{document}